\documentclass{egpubl}
\usepackage{egsgp17}

\JournalSubmission    
\ifpdf \usepackage[pdftex]{graphicx} \pdfcompresslevel=9
\else \usepackage[dvips]{graphicx} \fi

\PrintedOrElectronic

\usepackage{t1enc,dfadobe}

\usepackage{egweblnk}
\usepackage{cite}
\usepackage{amsmath}
\usepackage{amssymb}
\usepackage{mathtools}
\usepackage[T1]{fontenc}
\usepackage{libertine}
\usepackage[style=base]{caption}

\newcommand {\wil}[1]{{\color{green}\textbf{W: #1}\normalfont}}
\newcommand {\paul}[1]{{\color{blue}\textbf{P: #1}\normalfont}}

\newcommand {\holger}[1]{{\color{cyan}\textbf{H: #1}\normalfont}}

\newcommand {\dcut}{\textsc{DepthCut}}

\title[\dcut$: $ Improved Depth Edge Estimation using Multiple Unreliable Channels]%
{\dcut: Improved Depth Edge Estimation \\ Using Multiple Unreliable Channels \vspace*{-.3in}}

\author[Paul Guerrero, Holger Winnem\"{o}ller, Wilmot Li \& Niloy J. Mitra]
{\parbox{\textwidth}{\centering Paul Guerrero$^1$ \hspace{10pt}  Holger Winnem\"{o}ller$^2$ \hspace{10pt} Wilmot Li$^2$ \hspace{10pt} Niloy J. Mitra$^{1}$
	}
	\\
	{\parbox{\textwidth}{\centering
			$^1$University College London \hspace{0.5in}
			$^2$Adobe Research
		} 
	}
}

\begin{document}


\teaser{ \vspace*{-.25in}
    \setlength{\abovecaptionskip}{10pt plus 0pt minus 0pt}
	\includegraphics[width=\linewidth]{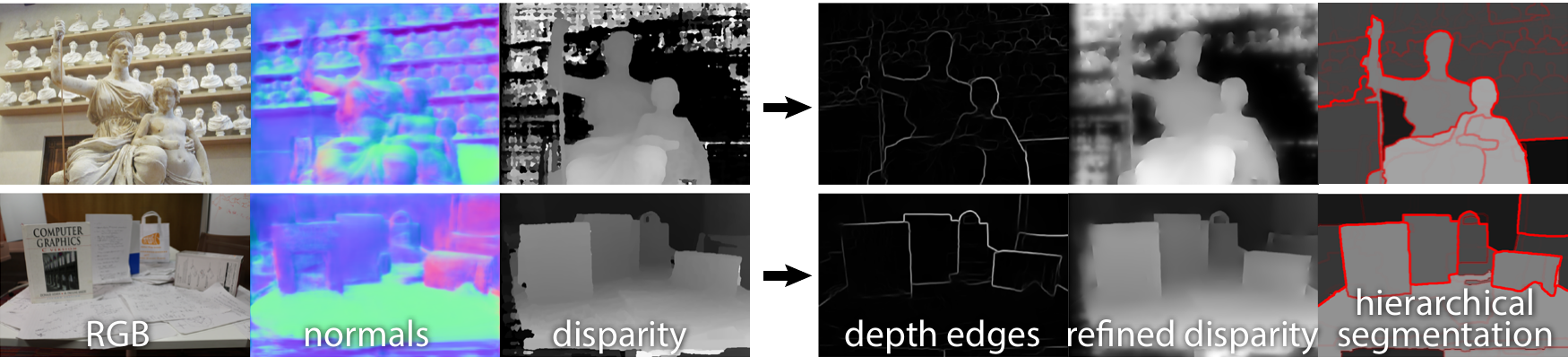}
	\centering
	\caption{We present \dcut, a method to estimate \emph{depth edges} with improved accuracy from unreliable input channels, namely: RGB images, normal estimates, and disparity estimates. Starting from a single image or pair of images, our method produces depth edges consisting of depth contours and creases, and separates regions of smoothly varying depth. Complementary information from the unreliable input channels are fused using a neural network trained on a dataset with known depth. The resulting depth edges can be used to refine a disparity estimate or to infer a hierarchical image segmentation.}
	\label{fig:teaser}
}

\maketitle

\begin{abstract}
\vspace{-5pt}	

In the context of scene understanding, a variety of methods exists to estimate different information channels from mono or stereo images, including disparity, depth, and normals. 
Although several advances have been reported in the recent years for these tasks, the  estimated information is often imprecise particularly near depth discontinuities or creases. %
Studies have however shown that precisely such \emph{depth edges} carry critical cues for the perception of shape, and play important roles in tasks like depth-based segmentation or foreground selection.
Unfortunately, the currently extracted channels often carry conflicting signals, making it difficult for subsequent applications to effectively use them. 
In this paper, we focus on the problem of obtaining high-precision depth edges (i.e., depth contours and creases) by jointly analyzing such unreliable information channels. 
We propose \dcut, a data-driven fusion of the channels using a convolutional neural network trained on a large dataset with known depth.
%
The resulting depth edges can be used for segmentation, decomposing a scene into depth layers with relatively flat depth, or improving the accuracy of the depth estimate near depth edges by constraining its gradients to agree with these edges.
Quantitatively, we compare against $15$ variants of baselines and demonstrate that our depth edges result in an improved segmentation performance and an improved depth estimate near depth edges compared to  data-agnostic channel fusion.
Qualitatively, we demonstrate that the depth edges result in superior segmentation and depth orderings. 
%
	
\end{abstract}


\section{Introduction}

\begin{figure*}[t]
    \setlength{\abovecaptionskip}{2pt plus 0pt minus 0pt}
	\includegraphics[width=1.0\textwidth]{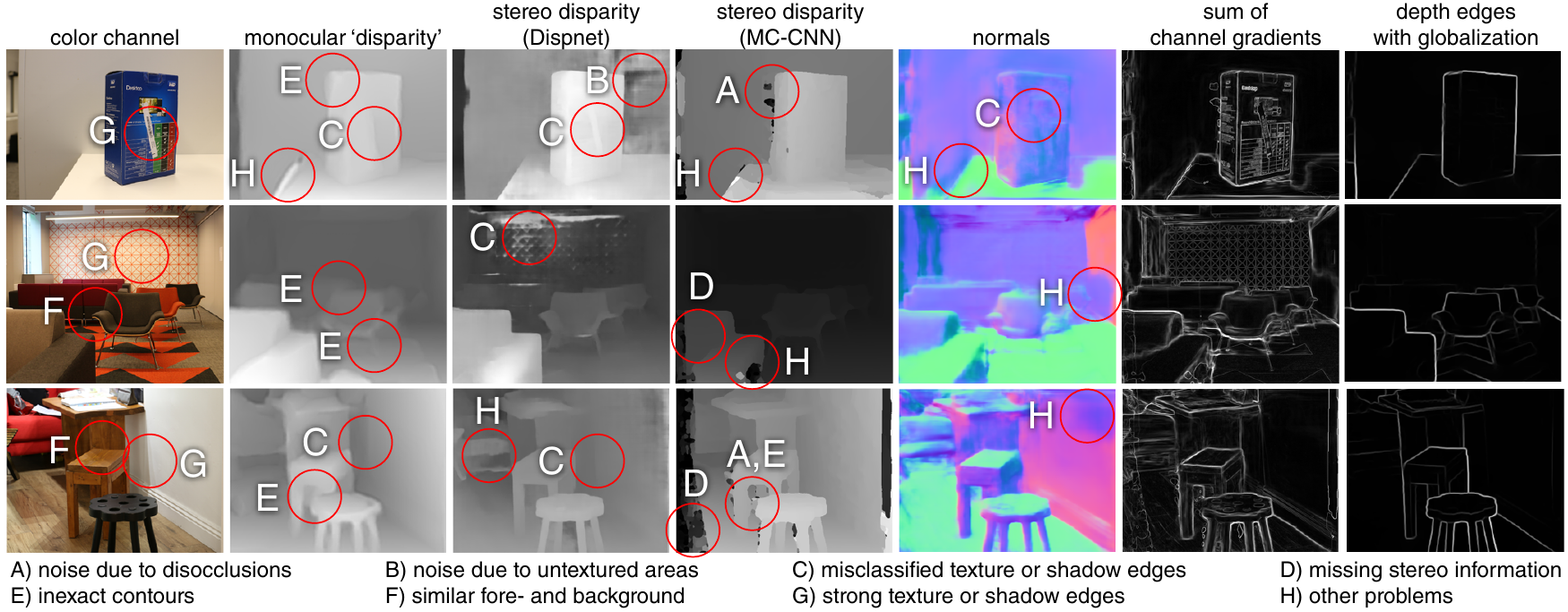}
	\caption{\footnotesize Unreliable input channels. The channels we use as input for depth edge estimation contain various sources of noise and errors. Error sources include areas of disocclusion, large untextured areas where stereo matching is difficult, and shadow edges that were incorrectly classified during creation of the channels. The color channel may also contain strong texture or shadow edges that have to be filtered out. The gradients of these channels do generally not align well, as shown in the second column from the right. We train \dcut\ to learn how to combine these channels (only including one of the disparity channels) to generate a cleaner set of depth edges, shown in the last column after a globalization step. In contrast to the sum of gradients, these depth edges now correspond to the probability of being a true depth contour or crease, giving them a larger intensity range. The optional globalization we show here only retains the most salient edges.
	}
	\label{fig:motivation}
\end{figure*}

A central task in scene understanding is to segment an input scene into objects and establish a (partial) depth-ordering among the detected objects. Since photographs remain the most convenient and ubiquitous option to capture scene information, a significant body of research has focused on scene analysis using single (mono) or pairs of (stereo) images. 
However, extracting high-quality information about scene geometry from such input remains a challenging problem.

Most recent mono and stereo scene estimation techniques attempt to compute disparity, depth or normals from the input image(s).
%
State-of-the-art methods largely take a data-driven approach by training different networks using synthetic (3D rendered) or other ground-truth data. Unfortunately, the resulting estimates still suffer from imperfections, particularly near depth discontinuities. Mono depth estimation is imprecise especially around object boundaries, while stereo depth estimation suffers from disocclusions and depends on the reliability of the stereo matching. Even depth scans (e.g., Kinect scans) have missing or inaccurate depth values near depth discontinuity edges. 

In this work, instead of aiming for precise depth estimates, we focus on identifying depth discontinuities, which we refer to as {\em depth edges}.  Studies (see Chapter 10 in \cite{gibson96} and \cite{Bansal:2013}) have shown that precisely such depth edges carry critical cues for the perception of shapes, and play important roles in tasks like depth-based segmentation or foreground selection. 
Due to the aforementioned artifacts around depth discontinuities, current methods mostly produce poor depth edges, as shown in Figure~\ref{fig:motivation}. 
Our main insight is that we can obtain better depth edges by fusing together multiple cues, each of which may, in isolation, be unreliable due to misaligned features, errors, and noise.
In other words, in contrast to absolute depth, depth edges often correlate with edges in other channels, allowing information from such channels to improve global estimation of depth edge locations.

We propose a data-driven fusion of the channels using \dcut, a convolutional neural network (CNN) trained on a large dataset with known depth. 
Starting from either mono or stereo images, we investigate fusing three different channels: color, estimated disparity, and estimated normals (see Figure~\ref{fig:motivation}).
The color channel carries good edge information wherever there are color differences. However, it fails to differentiate between depth and texture edges, or to detect depth edges if adjacent foreground and background colors are similar. 
Depth disparity, estimated from stereo or mono inputs, tends to be more reliable in regions away from depth edges and hence can be used to identify texture edges picked up from the color channel. It is, however, unreliable near depth edges as it suffers from disocclusion ambiguity. 
Normals, estimated from left image (for stereo input) or mono input, can help identify large changes in surface orientation, but they can be polluted by misclassified textures, etc. 
Combining these channels is challenging, since different locations on the image plane require different combinations, depending on their context. Additionally, it is hard to formulate explicit rules how to combine channels.
We designed \dcut\ to combine these unreliable channels to obtain robust depth edges. 
The network fuses multiple depth cues in a context-sensitive manner by learning what channels to rely on in different parts of the scene.
For example, in Figure~\ref{fig:teaser}-top,  \dcut\   correctly obtains depth segment layers separating the front statues from the background ones even though they have very similar color profiles; while in Figure~\ref{fig:teaser}-bottom, \dcut\ correctly segments the book from the clutter of similarly colored papers. In both examples, the network produces good results even though the individual channels are noisy due to color similarity, texture and shading ambiguity, and poor disparity estimates around object boundaries.

We use the  extracted depth edges for segmentation, decomposing a scene into depth layers with relatively flat depth, or improving the accuracy of the depth estimate near depth edges by constraining its gradients to agree with the estimated (depth) edges.
We extensively evaluate the proposed estimation framework, both qualitatively and quantitatively, and report consistent improvement over state-of-the-art alternatives. 
Qualitatively, our results demonstrate clear improvements in interactive depth-based object selection tasks on various challenging images (without available ground-truth for evaluation). We also show how \dcut\ can produce qualitatively better disparity estimates near depth edges.  
From a quantitative perspective, our depth edges lead to large improvements in segmentation performance compared to $15$ variants of baselines that either use a single channel or perform data-agnostic channel fusion. 
On a manually-captured and segmented test dataset of natural images, our \dcut-based method achieves an $f1$ score of $0.78$, which outperforms all $15$ variants of the baseline techniques by at least $9\%$.

%

\section{Related Work}

\paragraph*{Shape analysis.}
In the context of scene understanding, a large body of work focuses on estimating attributes for indoor scenes by computing high-level object features and analyzing inter-object relations (see \cite{Mitra:2014} for a survey). 
More recently, with renewed interest in convolutional neural networks, researchers have explored data-driven approaches for various shape and scene analysis tasks (cf., \cite{Xu:2016}). 
While there are too many efforts to list, representative examples include normal estimation~\cite{NNdepth,normalEst:16}, object detection~\cite{Szegedy2013DeepNN}, semantic segmentation~\cite{Couprie2013IndoorSS,Gupta2014LearningRF}, localization~\cite{Sermanet2014IntegratedRL}, pose estimation~\cite{Toshev2014DeepPoseHP,Bakry2015DiggingDI}, and scene recognition using combined depth and image features from RGBD input~\cite{Zhu2016DiscriminativeMF}, etc.

At a coarse level, these data-driven approaches produce impressive results, but they are often noisy near discontinuities and in areas of fine detail. Moreover, the various methods tend to produce different types of errors in regions of ambiguity. Since each network is trained independently, it is hard to directly fuse the different estimated quantities (e.g., disparity and normals) to produce higher quality results.
 Finally, the above networks are largely trained on indoor scene datasets (e.g., NYU dataset) and do not usually generalize to new types of objects. 
Such limitations reduce the utility of these techniques in applications like segmentation into depth-ordered layers or disparity refinement, which require clean, accurate depth edges.
Our data-driven approach is to jointly learn the error correlations across different channels in order to produce high quality depth edges from mono or stereo input.

\paragraph*{General segmentation.}

In the context of non-semantic segmentation (i.e., object-level region extraction without assigning semantic labels), one of the most widely used interactive segmentation approaches is GrabCut~\cite{Rother:2004:GIF}, which builds GMM-based foreground and background color models.  
The state-of-the-art in non-semantic segmentation is arguably the method of Arbel\'{a}ez et al.~\cite{Arbelaez:2011}, which operates at the level of contours and yields a hierarchy of segments. 
Classical segmentation methods that target standard color images have also been extended to make use of additional information.
For example, Kolmogorov et al.~\cite{Kolmogorov:2005} propose a version of GrabCut that handles binocular stereo video, Sundberg et al.~\cite{Sundberg:2011} compute depth-ordered segmentations using optical flow from video sequences, and Dahan et al.~\cite{Dahan2012} leverage scanned depth information to decompose images into layers. 
In this vein, \dcut\ leverages additional channels of information (disparity and normals) that can be directly estimated from input mono or stereo images. By doing so, our method performs well even in ambiguous regions, such as textured or shaded areas, or where foreground-background colors are very similar. In Section~\ref{sec:results}, we present various comparisons with state-of-the-art methods and their variants.

\paragraph*{Layering.} 
Decomposing visual content into a stack of overlapping layers produces a simple and flexible ``2.1D'' representation~\cite{Nitzberg:1990} that supports a variety of interactive editing operations, such as those described in~\cite{McCann:2009}.
Previous work explores various approaches for extracting 2.1D representations from input images.
Amer et al.~\cite{Amer:2010} propose a quadratic optimization that takes in image edges and T-junctions to produce a layered result, and later generalize the formulation using convex optimization. More recently, Yu et al.~\cite{Yu:2014} propose a global energy optimization approach. 
Chen et al.~\cite{Chen:2013} identify five different occlusion cues (semantic, position, compactness, shared boundary, and junction cues) and suggest a preference function to combine these cues to produce a 2.1D scene representation. 
Given the difficulty of extracting layers from complex scenes, interactive techniques have also been proposed~\cite{Iizuka:2014}.
We offer an automatic approach that combines color, disparity and normal information to decompose input images into layers with relatively flat depth.

%

\section{Overview}

\begin{figure}[t!]
	\includegraphics[width=1.0\columnwidth]{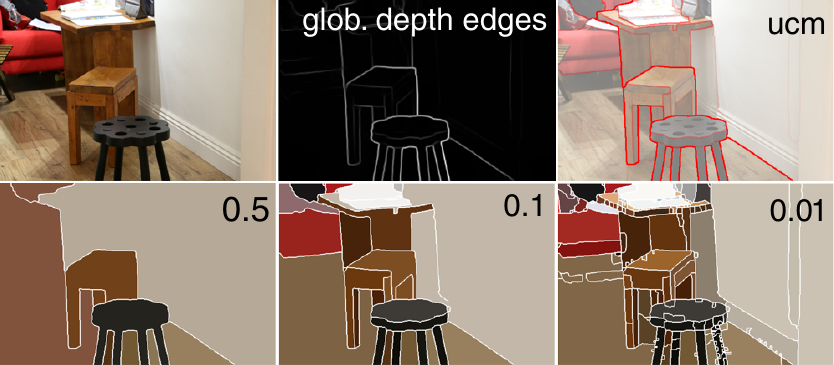}
	\caption{\footnotesize Example of a region hierarchy obtained using depth edges estimated by \dcut. The cophenetic distance between adjacent regions (the threshold above which the regions are merged) is based on the strength of depth edges. The \emph{Ultrametric Contour Map}~\cite{Arbelaez:2011} shows the boundaries of regions with strength proportional to the cophenetic distance. Thresholding the hierarchy yields a concrete segmentation, we show three examples in the bottom row.}
	\label{fig:hierarchy}
\end{figure}

\begin{figure*}[t!]
	\includegraphics[width=1.0\textwidth]{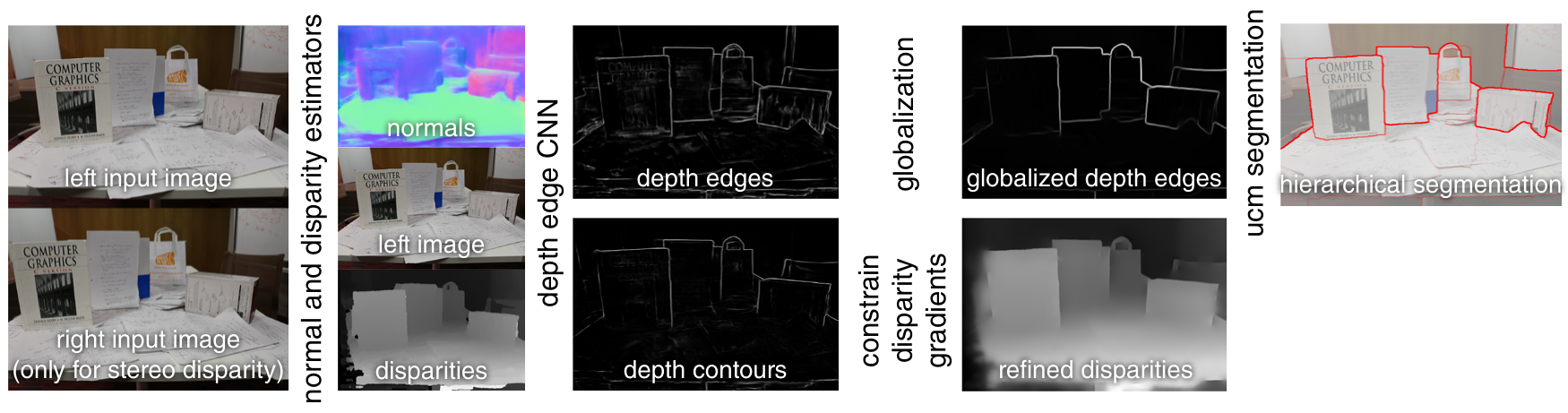}
	\caption{\footnotesize Overview of our method and two applications. Starting from a stereo image pair, or a single image for monocular disparity estimation, we estimate our three input channels using any existing method for normal or disparity estimation. These channels are combined in a data-driven fusion using our CNN to get a set of depth edges. These are used in two applications, segmentation and refinement of the estimated disparity. For segmentation, we perform a globalization step that keeps only the most consistent contours, followed by the construction of a hierarchical segmentation using the gPb-ucm framework~\cite{Arbelaez:2011}. For refinement, we use depth contours only (not creases) and use them to constrain the disparity gradients.
	}
	\label{fig:overview}
\end{figure*}


\dcut\ estimates depth edges from either a stereo image pair or from a single image. Depth edges consist of depth contours and creases that border regions of smoothly varying depth in the image. They correspond to approximate depth- or depth gradient discontinuities. These edges can be used to refine an initial disparity estimate, by constraining its gradients based on the depth edges, or to segment an image into a hierarchy of
regions, giving us a depth layering of an image.
%
Regions higher up in the segmentation hierarchy are separated by stronger depth edges than regions further down, as illustrated in Figure \ref{fig:hierarchy}.

Given an accurate disparity and normal estimate, depth edges can be found based on derivatives of the estimates over the image plane.
In practice, however, such estimates are too unreliable to use directly (see Figures~\ref{fig:motivation} and \ref{fig:depthedges}). 
Instead, we fuse multiple unreliable channels to get a more accurate estimate of the depth edges. Our cues are the left input image, as well as a disparity and normal estimate obtained from the input images. These channels work well in practice, although additional input channels can be added as needed.
In the raw form, the input cues are usually inconsistent, i.e., the same edge, for example, may be present at different locations across the channels, or the estimates may contain edges that go missing in the other channels due to estimation errors.

The challenge then lies in fusing these different unreliable cues to get a consistent set of depth edges. The reliability of such channel features at a given image location may depend on the local context of the cue. For example, the color channel may provide reliable locations for contour edges of untextured objects, but may also contain unwanted texture and shadow edges. The disparity estimate may be reliable in highly textured regions, but  inaccurate at disocclusions. 
Instead of hand-authoring rules to combine such conflicting channels, we train a convolutional neural network~(CNN) to provide this context-sensitive fusion, as detailed in Section~\ref{sec:depth_edges}.

The estimated depth edges may be noisy and are not necessarily closed. To get a clean set of closed contours that decompose the image into a set of 2.1D regions, we adapt the non-semantic segmentation method proposed by Arbel\'{a}ez et al.~\cite{Arbelaez:2011} (see Figure~\ref{fig:results_segmentation_qual} to compare result of using the method directly on the individual channels or their naive combinations).
Details are provided in Section~\ref{sec:segmentation}.
The individual steps of our method are summarized in Figure~\ref{fig:overview}.




\section{Depth Edges}
\label{sec:depth_edges}


\begin{figure}[b!]
	\includegraphics[width=1.0\columnwidth]{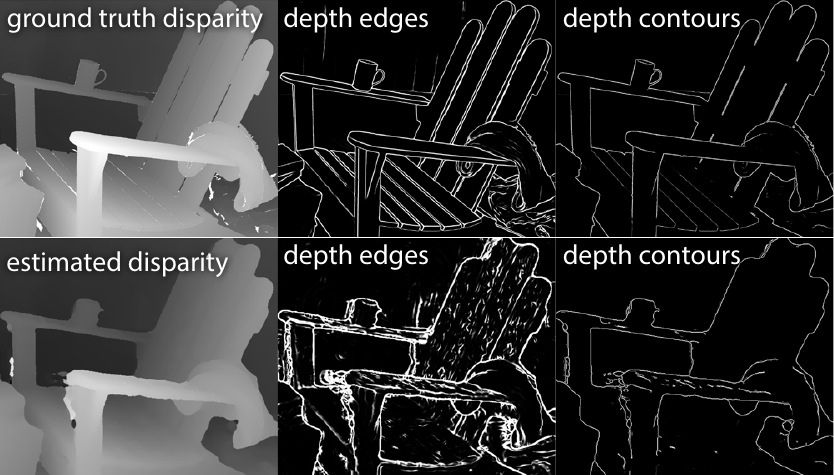}
	\caption{\footnotesize Depth edges and contours computed by applying the definition directly to ground-truth disparities (top row) and estimated disparities (bottom row). The high-order terms in the definition result in very noisy edges for the disparity estimate.}
	\label{fig:depthedges}
\end{figure}

Depth edges consist of depth contours and creases. These edges separate regions of smoothly varying depth in the image, which can be used as segments, or to refine a disparity estimate. Our goal is to robustly estimate these depth edges from either a stereo image pair or a single image.

We start with a more formal definition of depth edges. Given a disparity image as continuous function $D(u,v)$ over locations $(u,v)$ on the image plane, a \emph{depth contour} is defined as a $C^0$ discontinuity of $D$. In our discrete setting,  however, it is harder to identify such discontinuities. Even large disparity gradients are not always reliable as they are also frequently caused by surfaces viewed at oblique angles. Instead, we define the probability $P_c$ of {\em depth contour} on the positive part of the Laplacian of the gradient:
\begin{equation*}
P_c(u,v) := \sigma_{\alpha} \bigl((\Delta \lVert \nabla D \rVert)^+ (u,v) \bigr),
\end{equation*}
where $\lVert \nabla D \rVert$ is the gradient magnitude of $D$, $\Delta$ is the Laplace operator, $(f)^+$ denotes the positive part of a function, and $\sigma$ is a sigmoid function centered at $\alpha$ that defines a threshold for discontinuities. We chose a logistic function $\sigma_\alpha(x) = 1 / (1+e^{-10(x/\alpha-1)})$ with a parameter $\alpha = 1$.

Creases of 3D objects are typically defined as strong maxima of surface curvature. However, we require a different definition, since we want our creases to be invariant to the scale ambiguity of objects in images; objects that have the same appearance in an image should have the same depth creases, regardless of their world-space size. We therefore take the normal gradient of each component of the normal separately over the \emph{image plane} instead of the divergence over geometry surfaces.
Given a normal image $N(u,v) \in \mathbb{R}^3$ over the image plane, we define the probability $P_r$ of \emph{depth creases} on gradient magnitude of each normal component:
\begin{equation*}
P_r(u,v) := \sigma_{\beta} \bigl( (\|\nabla N_x\|+\|\nabla N_y\|+\|\nabla N_z\|) (u,v) \bigr),
\end{equation*}
where $N_x$, $N_y$ and $N_z$ are the components of the normal,
and $\sigma$ is the logistic function centered at $\beta = 0.5$. The combined probability for a \emph{depth edge} $P_e(u,v)$ is then given as: 
\begin{equation*}
P_e(u,v) := \bigl( 1-(1-P_c) (1-P_r) \bigr) (u,v).
\end{equation*}
This definition can be computed directly on reliable disparity and normal estimates. For unreliable and noisy estimates, however, the high-order derivatives amplify the errors, examples are shown in Figure~\ref{fig:depthedges}. In the next section, we discuss how \dcut\ estimates the depth edges using  unreliable disparity and normals.



\section{Depth Edge Estimation}

\begin{figure*}[t!]
	\includegraphics[width=1.0\textwidth]{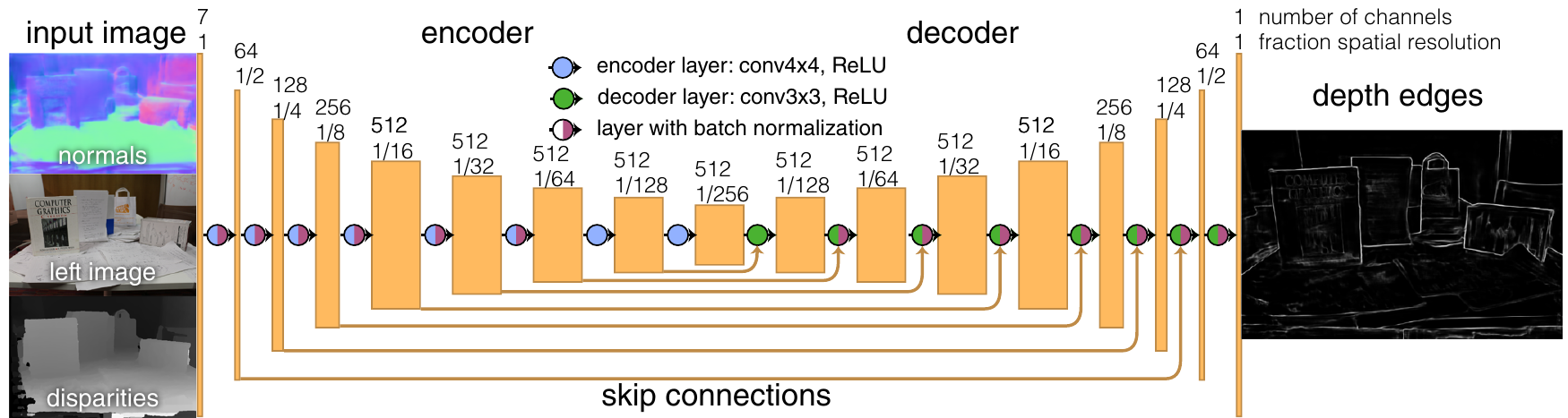}
	\caption{\footnotesize CNN architecture for depth edge estimation. The orange boxes are layer input/output multi-channel images, colored disks are layers. Starting from a set of input channels, the encoder extracts a set of features of increasing abstraction while downsampling the feature maps. The decoder uses these abstracted features to construct the depth edges. Skip connections let information flow around the bottleneck between encoder and decoder.}
	\label{fig:net}
\end{figure*}

We obtain disparity and normal estimates by applying state-of-the-art estimators either to the stereo image pair, or to the left image only. Any existing stereo or mono disparity, and normal estimation method can be used in this step. Later, in Section~\ref{sec:results}, we report performance using various disparity estimation methods.

The estimated disparity and normals are usually noisy and contain numerous errors. A few typical examples are shown in Figure~\ref{fig:motivation}. The color channel
is more reliable, but contains several other types of edges as well, such as texture and shadow edges. By iteself, the color channel alone provides insufficient information to distinguish between depth edges and these unwanted types of edges.

Our key insight is that the reliability of individual channels at each location in the image can be estimated from the full set of channels. 
For example, a short color edge at a location without depth or normal edges is likely to be a texture edge, or edges close to a disocclusion are likely to be noise if there is no evidence for an edge in the other channels. It would be hard to formulate explicit rules for these statistical properties, especially since they may be dependent on the specific estimator used.
This motivates a data-driven fusion of channels, where we avoid hand-crafting explicit rules in favor of training a convolutional neural network to learn these properties from data. We will show that this approach gives better depth edges than a data-agnostic fusion. 

\subsection{Model}
Deep neural networks have a large capacity to learn a context-sensitive fusion of channels based on different cues for their local reliability. We use a convolutional neural network (CNN), a type of network that has shown remarkable performance on a large range of image processing tasks~\cite{Krizhevsky:2012, Simonyan:2015}.
The image is processed through a series of successive non-linear filter banks called layers, each operating on the output of the previous layer. The output of each layer is a multi-channel image $x \in \mathbb{R}^{w \times h \times c}$, where $w$ and $h$ are the width and height of the image and $c$ corresponds to the number of filters in the layer:
\begin{equation*}
    \mathcal{L}\ :\ \mathbb{R}^{w\ \times\ h\ \times\ c} \rightarrow \mathbb{R}^{w' \times\ h' \times\ c'}.
\end{equation*}
Each output channel can be understood as a \emph{feature map} extracted from the input by one of the filters. The input to the first layer is composed of the RGB channels $I$ of the input image with size $W \times H \times 3$, the disparity estimate $\widetilde{D}$, and the $xyz$ channels of the normal estimate  $\widetilde{N}$, giving a total size of $W \times H \times 7$. The output of the last layer is the estimated probability $\widetilde{P}_e$ for a depth edge over the image:
\begin{equation}
\widetilde{P}_e := \bigl(\mathcal{L}_n(p_n) \circ \dots \circ \mathcal{L}_2(p_2) \circ \mathcal{L}_1(p_1)\bigr) (X),
\end{equation}
where $X$ is the concatenation of $I$, $\widetilde{D}$, and $\widetilde{N}$ into a single multi-channel image and $p_i$ are the parameters of the  i-th layer.
Optimization of the model parameters is based on gradient descent, so each $\mathcal{L}$ must be differentiable, or at least subderivatives must exist.
%

\paragraph*{Layers.}
Each layer convolves its input with the filter bank of the layer and adds a bias, before applying a non-linear \emph{activation function} $\sigma: \mathbb{R} \rightarrow \mathbb{R}$, that gives the model its non-linearity:
\begin{equation*}
    \mathcal{L}^j(x\ |\ w^j,b^j) =  \sigma\bigl((x\ \ast_{UV}\ w^j) + b^j\bigr),
\end{equation*}
where $\mathcal{L}^j$ denotes a single output channel $j \in [1,c']$ of $\mathcal{L}$, that is, a single feature map. The parameters $p_i$ of the entire layer consist of one kernel $w^j \in \mathbb{R}^{k_w \times k_h \times c}$ and one bias $b^j \in \mathbb{R}$ per feature map.
Each kernel spans all input channels and the convolution is over the spatial domain $(u,v) \in [1,w] \times [1,h]$ only, denoted by $\ast_{UV}$.
For the activation function, a typical choice is the Rectified Linear Unit (ReLU), that clamps inputs to values above $0$. He et al.~\shortcite{He:2015} show that `leaky' versions of the ReLU that have small but non-zero derivatives for negative values perform better, since  the vanishing derivative of the original ReLU can become problematic for gradient descent. We use a small constant derivative for negative values to avoid introducing additional parameters to the model:
\begin{equation*}
\sigma_{\mathrm{LReLU}}(x) =
\begin{cases}
    x,     & \text{if } x\geq 0\\
    0.2 x, & \text{otherwise.}
\end{cases}
\end{equation*}

One difficulty in optimizing this type of model is that parameters in later layers need to be adapted to changes in earlier layers, even if later layers already have optimal parameters. Ioffe et al.~\shortcite{Ioffe:2015} propose \emph{Batch Normalization} to address this problem. The output of a layer is normalized to zero mean and unit standard deviation, stabilizing the input for following layers. A layer is then defined as
\begin{equation*}
    \mathcal{L}^j(x\ |\ w^j,b^j) =  (\sigma \circ bn) \bigl((x\ \ast_{UV}\ w^j) + b^j\bigr),
\end{equation*}
where $bn$ denotes batch normalization.

\paragraph*{Encoder.}
We base our network on the encoder-decoder architecture~\cite{Hinton:2006,pix2pix2016}, which has been applied successfully to many image processing problems.
In a network of $n$ layers, the first $n/2$ layers act as an encoder, where consecutive layers extract features of increasing abstraction from the input. The remaining $n/2$ layers act as decoder, where the features are used to construct the depth edge probability image. Figure~\ref{fig:net} illustrates the architecture. The encoder layers progressively downsample the input, while increasing the number of channels to hold a larger number of features. An encoder layer is defined as:
\begin{equation*}
    \mathcal{E}^j(x\ |\ w^j,b^j) =  \curlyvee_2 \bigl(\mathcal{L}^j(x\ |\ w^j,b^j)\bigr),
\end{equation*}
where $\curlyvee_n$ denotes subsampling by a factor of $n$. We use a factor of $2$ for all our encoder layers. In practice this subsampling is implemented by increasing the stride of the convolution to avoid computing outputs for pixels that are discarded later on.

The spatial extent of filter kernel $w_i$, the downsampling factor and the number of preceding layers determine the receptive field size of the features in each layer. The receptive field is the neighborhood in the image that influences the computation of a feature; larger receptive fields can capture more global properties of an image. In general it is preferable to use smaller kernel sizes and deeper layers to achieve the same receptive field size: two consecutive $3 \times 3$ convolution layers are more expressive than a single $5 \times 5$ convolution layer, due to the extra non-linearity, while only needing $18$ parameters compared to $25$. However, this comes at an increased memory cost for storing the state of the additional layers. \dcut\ comprises of $8$ encoder layers, each with a kernel size of $4 \times 4$, for a maximum receptive field size of $256 \times 256$ pixels in the last encoder layer.

\paragraph*{Decoder.}
A decoder layer upsamples the input:
\begin{equation*}
    \mathcal{D}^j(x\ |\ w^j,b^j) =  \mathcal{L}^j(\curlywedge_2(x)\ |\ w^j,b^j),
\end{equation*}
where $\curlywedge_n$ denotes upsampling by a factor of $n$. We set the upsampling factor equal to the subsampling factor of the encoder layers, so that chaining an equal amount of encoder and decoder layers results in an output of the same size. An often employed alternative to upsampling is to replace the convolution with a \emph{deconvolution}~\cite{Shi:2016, Dumoulin:2016}, which transposes the convolution interpreted as a matrix. However, recent work~\cite{Odena:2016} suggests that using nearest-neighbor upsampling is preferable to deconvolutions, as it reduces visible checkerboarding artifacts.
The last layer of the decoder replaces the Leaky ReLU activation function with a sigmoid function to clamp the output to the $[0,1]$ range of depth edge probabilities.

\paragraph*{Information bottleneck.}
Our depth edge output has the same spatial resolution as the input, therefore depth edges need to be aligned to the fine details of the input channels. However, in the encoder-decoder architecture, the decoder operates only with the features in the last encoder layer. The output of this layer has the smallest spatial resolution in the network and is a bottleneck for information passing through the network. To circumvent this bottleneck and make use of all the features in the encoder, we use skip connections~\cite{Ronneberger:2015} that directly connect intermediate encoder- and decoder layers. Specifically, layers with the same resolution are connected, taking the output of the encoder layer before applying the activation function and concatenating it as additional channels to the output of the corresponding decoder layer before the activation function.

In our architecture, the minimum width and height of the input is $256 \times 256$ pixels. At this size, the feature maps in the output of the encoder have a spatial extent of a single pixel. Note that we do not use Batch Normalization in layers close to this bottleneck, since this would prohibit using small batch sizes. If the batch contains few pixels, it could potentially destroy a large part of the information contained in the batch. Batch Normalization removes two degrees of freedom from the batch (the mean and variance). For example 4-pixel batches (batch size 4 with a $256 \times 256$ input), would lose half of the information, and a single or two-pixel batch would lose all information.
For details, please refer to Figure~\ref{fig:net}.



\subsection{Loss and Training}
We trained our model by comparing our output to ground-truth depth edges. We experimented with various loss functions, but found the mean squared error to work best. We did find, however, that comparisons in our datasets were biased to contain more loss from false negatives due to errors or inaccuracies of the disparity estimate (i.e., ground-truth depth edges that were missing in the output because they were not present in the disparity estimate), than false positives due to texture or shadow edges (i.e., depth edges in the output due to texture or shadow edges that are not in the ground-truth). To counteract this bias, we multiply the loss at color channel edges that do {\em not} coincide with depth edges by a factor of $10$. Thus, we have 
\begin{equation*}
E(\widetilde{P}_e,P_e,M) = \frac{1}{n} \|M \odot (\widetilde{P}_e - P_e)\|^2_{\mathrm{FRO}},
\end{equation*}
where $\widetilde{P}_e$ and $P_e$ are the estimated and ground-truth depth edges, respectively, $M$ is a mask that takes on the value $10$ at color edges that do not coincide with depth edges and $1$ everywhere else, $\odot$ denotes element-wise multiplication, and $\|X\|^2_{\mathrm{FRO}}$ is the squared Frobenius norm of $X$, which sums up the weighted squared error of all $n$ elements in the output channels.

\begin{figure}[b!]
	\includegraphics[width=1.0\columnwidth]{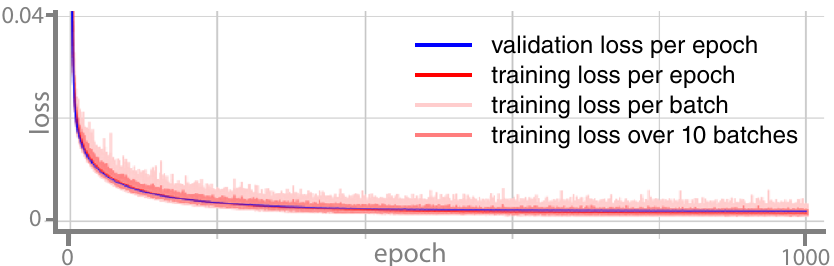}
	\caption{\footnotesize Typical loss curve when training our model. Notice that the validation and training loss are nearly identical, suggesting little overfitting to the training set.}
	\label{fig:training}
\end{figure}




To train the weights of the model, we use the Adam optimizer~\cite{Kingma:2015}, which resulted in faster training times than the more established Nesterov momentum~\cite{Nesterov:2005}, although both optimizers were found to converge to roughly the same solution.
To combat overfitting, we randomly sample patches of size $256 \times 256$ from the input images during training and add an $L_2$ regularization term $\lambda \|p\|_2^2$ to the loss,
where $p$ are the parameters of our model and the scaling  $\lambda$ set to $10^{-5}$, which effectively eliminates overfitting on our validation set. See Figure~\ref{fig:training} for a typical loss curve. In our experiments, we trained with a batch size of $5$ input patches.
For high resolution images, the receptive field of our network only covers a relatively small fraction of the image, giving our model less global information to work with. To decrease the dependence of our network on image resolution, we downsample high-resolution images to  $800$ pixel width while maintaining the original aspect ratio.

\section{Segmentation}
\label{sec:segmentation}
Since depth edges estimated by \dcut\ separate regions of smoothly varying depth in the image, as a first application we use it towards improved segmentation. Motivated by studies linking depth edges to perception of shapes, it seems plausible that regions divided by depth edges typically comprise simple shapes, that is, shapes that can be understood from the boundary edges only. Intuitively, our segmentation can then be seen as an approximate decomposition of the scene into simple shapes.

The output of our network typically contains a few small segments that clutter the image (see Figure~\ref{fig:net}, for example). This clutter is removed in a globalization stage, where global information is used to connect boundary segments that form longer boundaries and remove the remaining segments.

%
%
%

To construct a segmentation from these globalized depth edges, we connect edge segments to form closed contours. The OWT-UCM framework introduced by Arbel\'{a}ez et  al.~\shortcite{Arbelaez:2011} takes a set of contour edges and creates a hierarchical segmentation, based on an oriented version of the watershed transform (OWT), followed by the computation of an \emph{Ultrametric Contour Map} (UCM). The UCM is the dual of a hierarchical segmentation; it consists of a set of closed contours with strength corresponding to the probability of being a true contour (please refer to the original paper for details). A concrete segmentation can be found by merging all regions separated by a contour with strength lower than a given threshold (see Figure~\ref{fig:hierarchy}). 

The resulting UCM correctly separates regions based on the strength of depth edges, i.e., the \dcut\ output is used to build the affinity matrix. However, we found it useful to include a term that encourages regions with smooth, low curvature boundaries. Specifically, we go through the contour hierarchy from coarse to fine, adding one contour segment at a time, and modify their strength based on how likely they are the continuation of an existing boundary. Segments that smoothly connect to an existing stronger segment are strengthened by the following factor:
\begin{equation*}
w'_i = 1 + \max_{\{j | w_j > w_i \}} \left|\cos(\angle_{ij})\right| \bigl(\max(1,0.5\frac{w_j}{w_i} \sqrt{\sigma(l_i) \sigma(l_j)}\bigr),
\end{equation*}
where $w_i$ is the current strength of a boundary segment and $w'_i$ is the updated strength. The maximum is over all segments $j$ in a coarser hierarchy level (with larger strength) that are connected to segment $i$. The angle $\angle_{ij}$ denotes the angle between the tangents of segments at their connection point. $l_i$ is the arc length of segment $i$, which intuitively biases the formulation towards long segments that are less likely to be clutter and $\sigma$ is a logistic function that saturates the arc length above a given threshold, we use $0.7$ times the image width in our experiments.


\section{Depth Refinement}
\label{sec:refinement}

As a second application of our method, we can refine our initial disparity estimates. For this application, we train our network to output \emph{depth contours} as opposed to depth edges, i.e., a subset of the depth edges. In addition to depth contours, we also train to output \emph{disparity gradient directions} as two normalized components $\widetilde{d}_u$ and $\widetilde{d}_v$. Due to our multi-channel fusion, the depth contours are usually considerably less noisy than the initial disparity estimate. They provide  more accurate spatial locations for strong disparity gradients, while $\widetilde{d}_u$ and $\widetilde{d}_v$ provide accurate orientations. However, we do not obtain  the actual gradient magnitudes. Note that getting an accurate estimate of this magnitude over the image would be a much harder problem, since it would require regressing the gradient magnitude instead of classifying the existence of a contour.

\begin{figure*}[t!]
	\includegraphics[width=1.0\textwidth]{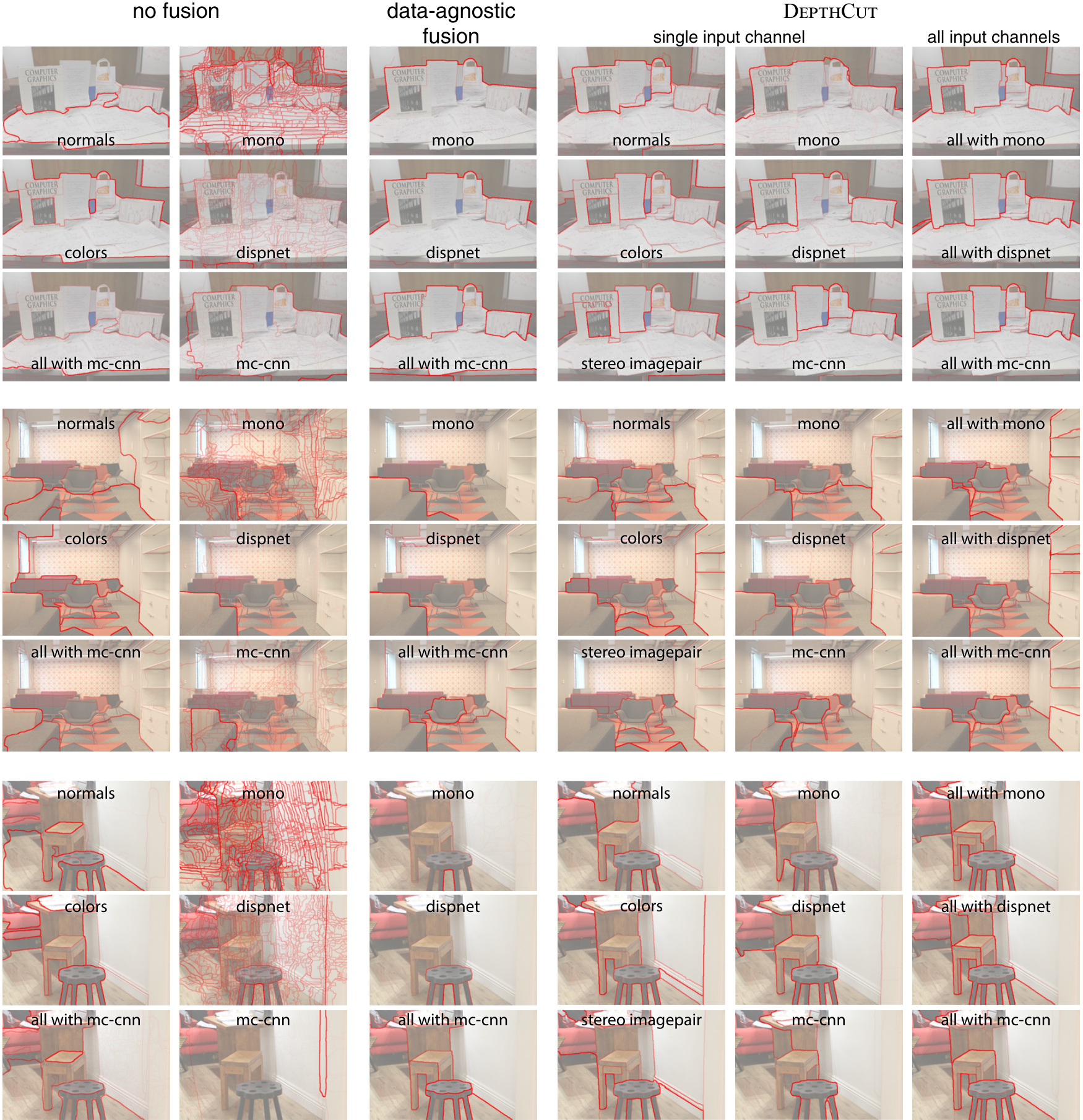}
	\caption{\footnotesize Hierarchical segmentations based on our depth edges. We compare directly segmenting the individual channels, performing a data-agnostic fusion, and applying our data driven fusion with either a subset of the input channels, or all of the input channels. Strongly textured regions in the color channel make finding a good segmentation difficult, while normal and disparity estimates are too unreliable to use exclusively. Using our data-driven fusion gives segmentations that better correspond to objects in the scenes.
	}
	\label{fig:results_segmentation_qual}
\end{figure*}

We obtain a smooth approximation of the gradient magnitude from the disparity estimate itself and only use the depth contours and disparity directions to decide if a location on the image plane should exhibit a strong gradient or not, and to constrain the gradient orientation.
In the presence of a depth edge, we constrain the dot product between the initial disparity gradient and the more reliable directions $\widetilde{d}_u$ and $\widetilde{d}_v$ to be above a minimum value. This minimum should be proportional to the actual disparity change in direction $(\widetilde{d}_u,\widetilde{d}_v)$ integrated over some fixed distance. Intuitively, we want to focus this disparity change at the edge location, while flattening the change to zero at locations without a depth edge.
This can be formulated as a linear least squares problem:
\begin{equation}
\begin{aligned}
& \underset{x}{\text{min}} & & \|\widetilde{P}_e^T (\widetilde{d}_u G_u x + \widetilde{d}_v G_v x - y - c)\|_2^2\\
& & +\ & \|(1-\widetilde{P}_e)^T G_u x \|_2^2 + \| (1-\widetilde{P}_e)^T G_v x\|_2^2\\
& & +\ & \mu \|x-x_0\|_2^2 \\
& \text{s.t.} & & y \geq 0,
\end{aligned}
\end{equation}
where $x$ are the disparities written as a vector, $G_u$ and $G_v$ are matrix formulations of the kernel for the gradient $u$ and $v$ components and $y$ are slack variables. The first term soft-constrains the dot product of the disparity gradient with direction $(\widetilde{d}_u,\widetilde{d}_v)$ to be above a minimum value $c$ at locations with a depth edge $\widetilde{P}_e$. The following two terms soft-constrain the gradient $u$ and $v$ components to be $0$ at locations without a depth edge. The last term is a regularization term to prefer results close to the original disparities $x_0$. We solve this optimization problem  using the reflective Newton method implemented in MATLAB~\cite{Coleman:1996}.
For faster convergence, we implemented a multi-scale approach that performs the optimization iteratively from coarsest to finest scale on a 3-level image pyramid, with a down-scaling factor of $2$.

\begin{figure*}[t!]
	\includegraphics[width=1.0\textwidth]{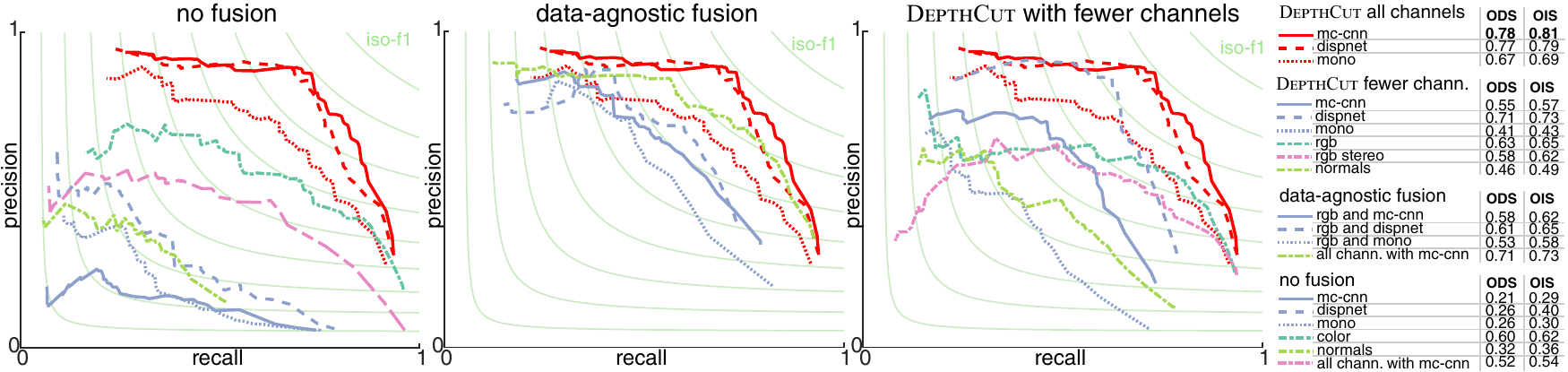}
	\caption{\footnotesize Quantitative comparison to all baselines on our Camera dataset. We show precision vs. recall over thresholds of the segmentation hierarchy. Note that depth edges from monocular disparity estimates (finely dotted lines) work with less information than the other two disparity estimates and are expected to perform worse. The depth edge estimates of our data-driven fusion, shown in red, consistently perform better than other estimates. The table on the right shows the f1 score for the best single threshold over the entire dataset (ODS), or the best threshold for each instance (OIS).}
	\label{fig:results_segmentation_quant}
\end{figure*}

\section{Results and Discussion}
\label{sec:results}

To evaluate the performance of our method, we compare against several baselines. 
We demonstrate that fusing multiple channels results in better depth edges than using single channels by comparing the results of our method when using all channels against our method using \emph{fewer channels} as input. (Note that the extra channels used in \dcut\ are estimated from only mono or stereo image inputs, i.e., all the methods have same source inputs to work with.) 
To support our claim that a data-driven fusion performs better than a \emph{data-agnostic} fusion, we compare to a baseline using manually defined fusion rules to compute depth edges. For this method, we use large fixed kernels to measure the disparity or normal gradient across image edges.
We also test providing the \emph{un-fused} channels directly as input to a segmentation, using the well-known gPb-ucm~\cite{Arbelaez:2011} method, the ucm part of which we use for our segmentation application, as well.

For each of these methods we experiment with different sets of input channels, including the color image, normals, and $3$ different disparity types from state-of-the-art estimators: the mc-cnn stereo matcher~\cite{Zbontar:2016}, dispnets~\cite{Mayer:2016} and a monocular depth estimate~\cite{Chakrabarti:2016}, for a total of $15$ baseline variations. More detailed results for our method are provided as supplementary materials, while here we present only the main results.

\subsection{Datasets}
We use two datasets to train our network, the Middlebury 2014 Stereo dataset~\cite{Scharstein:2014} and a custom synthetic indoor scenes dataset we call the \textsc{room}-dataset.
Even though the Middlebury dataset is non-synthetic, it has excellent depth quality. We cannot use standard RGBD dataset typically captured with noisy sensors like the Kinect, because the higher-order derivatives in our our depth edge definition are susceptible to noise (see Figure~\ref{fig:depthedges}). The Middlebury 2014 dataset consists of $23$ images of indoor scenes containing objects in various configurations, each of which was taken under several different lighting conditions and with different exposures. We perform data-augmentation by randomly selecting an image from among the exposures and lighting conditions during training and by randomizing the placement of the $256 \times 256$ patch in the image, as described earlier.

The room dataset consists of $132$ indoor scenes that were obtained by generating physically plausible renders of rooms in the scene synthesis dataset by Fisher et al.~\cite{Fisher:2012} using a light tracer. Since this is a synthetic dataset, we have access to perfect depth. 
Recently, several synthetic indoor scene datasets have been proposed~\cite{McCormac:2016,Song:2017} that would be good candidates to extend the training set of our method, we plan to explore this option in future work.
The ground-truth on these datasets is created by directly applying the depth edge definition in Section~\ref{sec:depth_edges} to the ground-truth disparity. Since the disparity in the Middlebury dataset still contains small amounts of noise, we compute contour edges with a relatively large filters to get smoothed estimates. Depth contours are computed using a derivative-of-Gaussian filter, followed by a difference-of-Gaussians filter. For depth crease estimation, we reconstruct the points cloud using the known camera parameters, estimate normals from surface tangents, and compute the gradient of each component with a derivative-of-Gaussian filter. To remove noise while preserving contours, we filter the image with a large $15 \times 15$ median filter after reconstructing the point cloud and after computing the normals. We will release this dataset and generation scripts for future use. 

Our network performs well on these two training datasets, as evidenced by the low validation loss shown in Figure~\ref{fig:training}, but to confirm the generality of our trained network, we tested the full set of baselines on an unrelated dataset of $8$ images (referred to as the \textsc{camera}-dataset)  taken manually under natural (non-studio) conditions, for a total of $120$ comparisons with all baseline methods. These images were taken with three different camera types: a smartphone, a DSLR camera, either hand-held or on a tripod, and a more compact handheld camera. They contain noise, blur and the stereo images are not perfectly aligned. For these images, it is difficult to get an accurate depth ground truth, so we generated ground-truth depth edges by manually editing edge images obtained from the color channel, removing texture- and shadow edges, as well as edges with depth contours or creases below a threshold, keeping only prominent depth edges vital to a good segmentation to express a preference towards these edges, and adding missing depth edges (see the supplementary materials for this ground truth). For a future larger dataset of real-world images, we could either use Mechanical Turk to generate the ground truth, or employ an accurate laser scanner; although the latter would make the capturing process slower, limiting the number of images we could generate.





\begin{figure*}[t!]
	\includegraphics[width=1.0\textwidth]{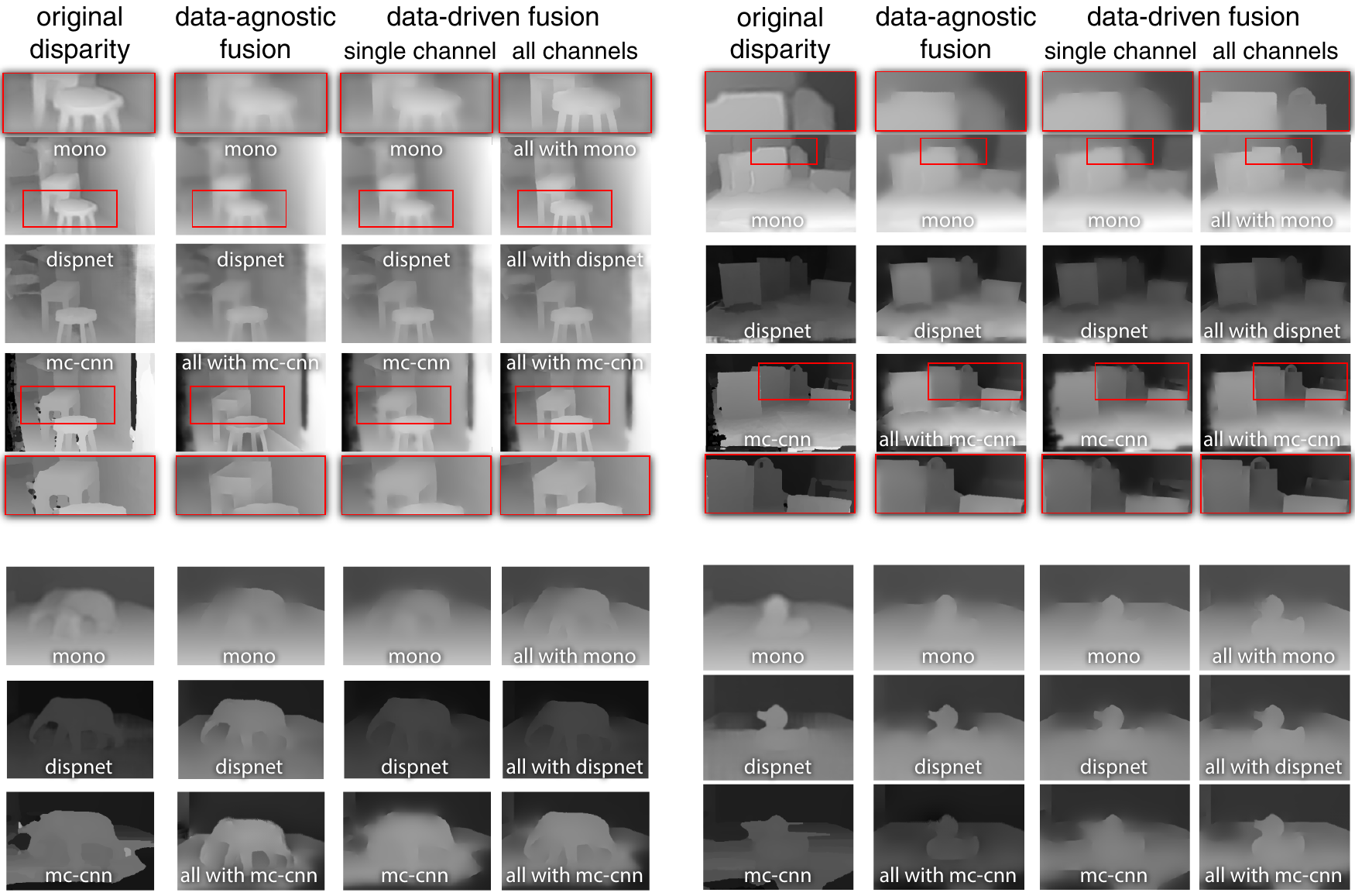}
	\caption{\footnotesize Estimated disparities refined with our depth edges. We compare our results to several other methods on four scenes. Our data-driven fusion reduces noise and enhances edges more robustly than methods either based on fewer channels or on data-agnostic fusion.}
	\label{fig:results_refinement_qual}
\end{figure*}

\subsection{Segmentation}

In the segmentation application, we compute a hieararchical segmentation over the image. This segmentation can be useful to select and extract objects from the image, or to composite image regions. Since we have approximate image depth, we also know the z-ordering of all regions, giving correct occlusions during editing.
For our segmentation application, we provide both qualitative and quantitative comparisons of the hierarchical segmentation on the \textsc{camera}-dataset.

\paragraph*{Qualitative comparisons.}
Figure~\ref{fig:results_segmentation_qual} shows the full set of comparisons on three images. For each image, the hierarchical segmentation of all $18$ methods (including our $3$ results) is shown in $3 \times 6$ tables. The large labels on top of the figure denote the method, while the smaller labels on top of the images denote the input channels used to create the image. Only the images that are labeled \emph{`all with \dots'} use multiple input channels, all other images were created with a single channel as input, and only the results that contain `dispnet', `'mc-cnn' or `stereo imagepair' in their label use a stereo image pair as input, the other results use a single image as input. Red lines show the UCM, stronger lines indicate a stronger separation between regions.

As is often the case in real-world photographs, these scenes contain a lot of strongly textures surfaces, making the objects in these scenes hard to segment without relying on additional channels. This is reflected in the methods based on color channel input that fail to consistently separate texture edges from depth edges. Another source of error are inaccuracies in the estimates, this is especially noticeable in the normal and monocular depth estimates, where the contours only very loosely follow the image objects. Using multiple channels without proper fusion does not necessarily improve the segmentation, as is especially evident in the multi-channel input of the un-fused method in lower-left corner of the $3 \times 6$ tables. \dcut\ can correct errors in the individual channels giving us region boundaries that are better aligned to depth edges, as shown in the right-most column.

\paragraph*{Quantitative comparisons.}
Quantitative tests were performed with all baselines images of the \textsc{camera}-dataset. We compare to the ground-truth using the Boundary Quality Metric~\cite{Martin:2004} that computes precision and recall of the boundary pixels. We use the less computationally expensive version of the metric, where a slack of fixed radius is added to the boundaries to not overly penalize small inaccuracies. Since we have hierarchical segmentations, we compute the metric for the full range of thresholds and report the value at the single best threshold over the entire dataset (ODS) or at the best value for each image (OIS).

Results are shown in Figure~\ref{fig:results_segmentation_quant}. The three plots show precision versus recall of the boundary pixel for all of the methods, averaged over all images. The $f1$ score, shown as iso-lines in green, summarizes precision and recall into a single statistic where higher values (towards the top-right in the plots) are better. Note that the faintly dotted lines correspond to monocular depth estimates that operate with less information than stereo estimates and are expected to perform worse. The table on the right-hand side show the ODS and OIS values for all methods. Note that our fusion generally performs best, only the monocular depth estimates have a lower score than the stereo estimates of some other methods.

\subsection{Disparity Refinement}

The second application for our method is disparity refinement, where we improve an initial disparity estimate based on our depth edges. Figure~\ref{fig:results_refinement_qual} shows results in a similar layout to Figure~\ref{fig:results_segmentation_qual} with zoomed insets in red to highlight differences. It is important to note that we only improve the disparity near depth edges, but these are often the regions where disparty estimates have the largest errors. For example, our refinement step can successfully reduce noise caused by disocclusions of a stereo matching method, sharpen the blurry outlines of monocular depth estimation, create a gradient between contours that were incorrectly merged to the background, like in the `duck' or the  `elephant' image.

\section{Conclusions}
We present a method that produces accurate depth edges from mono or stereo images by combining multiple unreliable information channels (RGB images, estimated disparity, and estimated normals).
The key insight is that the above channels, although noisy, suffer from different types of errors (e.g., in texture or depth discontinuity regions), and a suitable context-specific filter can fuse the information to yield high quality depth edges. 
To this end, we trained a CNN using ground-truth depth data to perform this multi-channel fusion, and our qualitative and quantitative evaluations show significant improvement over alternative methods. 

We see two broad directions for further exploration. From an analysis standpoint, we have shown that data-driven fusion can be effective for augmenting color information with estimated disparity and normals. One obvious next step is to try incorporating even more information, such as optical flow from input video sequences. While this imposes additional constraints on the capture process, it may help produce even higher quality results. Another possibility is to apply the general data-driven fusion approach to other image analysis problems beyond depth edge estimation. The key property to consider for potential new settings is that there should be good correlation between the various input channels. 

Another area for future research is in developing more techniques that leverage estimated depth edges. We demonstrate how such edges can be used to refine disparity maps and obtain a segmentation hierarchy with a partial depth-ordering between segments. 
While our work already demonstrates how such edges can be used to refine disparity maps, we feel there are opportunities to further improve depth and normal estimates. The main challenge is how to recover from large depth errors, as our depth edges only provide discontinuity locations rather than the gradient magnitudes.
It is also interesting to consider the range of editing scenarios that could benefit from high quality depth edges. For example, the emergence of dual camera setups in mobile phones raises the possibility of on-the-fly, depth-aware editing of captured images. In addition, it may be possible to support a class of pseudo-3D edits based on the depth edges and refined depth estimates within each segmented layer.

\bibliographystyle{eg-alpha-doi}
\bibliography{depthlayering}

\end{document}